\documentclass{esannV2}
\usepackage[dvips]{graphicx}
\usepackage[latin1]{inputenc}
\usepackage{amsmath}
\usepackage{amssymb}
\usepackage{amsthm}
\usepackage{color}
\usepackage{subcaption}
\usepackage{hyperref}

\newcommand{\Tensor}[1]{\ensuremath{\mathbf{ #1}}}
\graphicspath{{plots/}}

\begin{document}
\title{Tensor Decompositions in Recursive Neural Networks for Tree-Structured Data}
\author{Daniele Castellana and Davide Bacciu
%
\thanks{This work has been supported by MIUR under project SIR 2014 LIST-IT (RBSI14STDE).}
%
\vspace{.3cm}\\
%
Dipartimento di Informatica - Universit\`a di Pisa - Italy
%
}

\maketitle

\begin{abstract}
The paper introduces two new aggregation functions to encode structural knowledge from tree-structured data. They leverage the Canonical and Tensor-Train decompositions to yield expressive context aggregation while limiting the number of model parameters. Finally, we define two novel neural recursive models for trees leveraging such aggregation functions, and we test them on two tree classification tasks, showing the advantage of proposed models when tree outdegree increases. 
\end{abstract}

\section{Introduction}
In the last decade, an interest in tensors and their decompositions has emerged in the machine learning community. This interest is firstly motivated by the natural representation of multi-modal data using tensor structures such as RGB images, videos or signals. A tensor representation of data is useful to capture complex interactions among input features which disappear if the data are "flattened" \cite{Cichocki2015TensorAnalysis}. Unfortunately, models that can handle multi-way data suffer a "curse of dimensionality": their weight are tensors whose size grows exponentially with the tensor order; tensor decompositions play a fundamental role to make this approach feasible \cite{Cichocki2015TensorAnalysis}. Also, tensor decompositions have been successfully applied to deep-learning models (1) to theoretical and practical bound the expressive power of a specific architecture and (2) to reduce the number of parameters without a dramatic reduction in performance \cite{Cichocki2017TensorPerspectives}.

Nevertheless, less attention has been paid to the role tensors (and their decompositions) could play to model high-order interaction among multiple input vectors. Multivariate functions are fundamental in models for tree-structured data since the structural knowledge of a node is computed by applying an aggregation function on pieces of information in the node's neighbourhood (i.e. the context). The choice of a simple aggregation function can lead to sub-optimal results; to this end, more complex functions based on Tucker tensor decomposition \cite{Kolda2009} have been used successfully both in probabilistic \cite{Castellana2019} and neural \cite{Castellana2020ECAI} models for tree-structured data.

The aim of this paper is to introduce two new aggregations functions which take advantage of two known tensor decompositions, i.e. Canonical \cite{Kolda2009} and Tensor Train \cite{Oseledets2011}. Both decompositions require a number of parameters which does not grow exponentially with respect to the tree maximum output degree, in contrast to the Tucker decomposition which has an exponential dependence between its rank and tree maximum output degree. Hence, we define new tensor-based tree-LSTM models which leverage the proposed functions to compute the hidden representation of input data. Finally, we compare experimentally all models assessing the impact of the tree maximum output degree on their performance.

\section{Tree-LSTM with Tensor Decompositions}\label{sec:tree-LSTM}
The Tree-LSTM \cite{Tai2015} is a generalisation of the standard LSTM architecture to tree-structured data. As in the standard LSTM, each LSTM unit contains: the \emph{input gate}, the \emph{output gate}, a \emph{forget gate} for each child node, and the \emph{memory cell}. A generic Tree-LSTM cell is defined by the following equations \cite{Tai2015}:
\begin{equation}
    \begin{alignedat}{2}
        i_v &= \sigma\left(f^i(x_v, h_{v1},\dots,h_{vL})\right),& \qquad o_v &= \sigma\left(f^o(x_v, h_{v1},\dots,h_{vL})\right),\\
        u_v &= \sigma\left(f^u(x_v, h_{v1},\dots,h_{vL})\right),& \qquad \bar{f}_{vj} &= \sigma\left(W^f x_v + U^f_{j} h_{vj} + b_j^f\right),\\
        c_v &= i_v \odot u_v + \sum_{j=1}^{L} \bar{f}_{vj} \odot c_{vj},& \qquad h_v &= o_v \odot \text{tanh}(c_v);
    \end{alignedat}
    \label{eq:treeLSTM}
\end{equation}
where $h_v \in \mathbb{R}^c$ and $\{x_v, h_{v1},\dots,h_{vL}\}$ are the hidden state and the context of node $v$ respectively. The context comprises the node label $x_v$, and all children hidden states; $h_{vj}$ indicates the hidden state of the $j$-th child of $v$; $L$ is the tree maximum outdegree. The terms $i_v$, $o_v$, $c_v$ are the input gate, the output gate and the memory cell, respectively; $\bar{f}_{vj}$ is the forget gate associated to the $j$-th child node, while the term $u_v$ is the value that is usually used to update the hidden state in a classic Tree-RNN. Following \cite{Tai2015}, we assume that the forget gate $\bar{f}_{vj}$ depends only on hidden state $h_{vj}$ rather than on the whole context to reduce the number of model parameters. In the generic Tree-LSTM defined in \eqref{eq:treeLSTM}, the values for $i_v$, $o_v$, $u_v$ are obtained through a neural network which is composed of an \emph{aggregation} function $f$ and a non-linear activation function $\sigma$. The superscript of $f$ indicates that we use different parameters for each aggregation function. In the next following, we define different Tree-LSTM architectures based on different aggregation functions. For the sake of clarity, we omit the gate superscript. Moreover, we remove the visible label $x_v$ from aggregation function inputs since in our experiments we take into account visible labels only on leaf nodes (i.e. nodes with no context); let $v$ the index of a leaf node, its hidden state is given by $h_v = \sigma(w(x_v)+b)$. The label on internal nodes (which indicates an operation in our task) is used to select a different parameterisation for the aggregation function. In all models, parameters can be learned through back-propagation.

\subparagraph{Sum-LSTM} defines an aggregation function which assumes complete independence among context elements. In particular, $f$ is defined as:
\begin{equation}
    f(h_{v1},\dots,h_{vL}) = u_1(h_{v1}) + \dots + u_L(h_{vL}) + b,
    \label{eq:sumLSTM}
\end{equation}
where $u_1, \dots, u_L: \mathbb{R}^c \rightarrow \mathbb{R}^c$ are linear maps and $b$ is the bias vector. The number of parameters required by the sum aggregation function is $O(Lc^2)$. The definition of our Sum-LSTM is equivalent to the Nary-LSTM defined in \cite{Tai2015}: we rename it to emphasise the use of the sum as the aggregation function.

\subparagraph{Full-LSTM} uses a multi-affine map, defining a fully tensorial approach to embedding computation. The multi-affine map uses an \emph{augmented tensor} \cite{Castellana2020ECAI}:
\begin{multline}
    f(h_{v1},\dots,h_{vL}) = f^{\Tensor{T}}(h_{v1},\dots,h_{vL}) = \\
    =\sum_{j_1=1}^{c+1}\dots\sum_{j_L=1}^{c+1}\Tensor{T}(j_1, \dots,j_L, k)     \bar{h}_1(j_1)\dots \bar{h}_L(j_L)
    \label{eq:fullLSTM},
\end{multline}
where $\Tensor{T} \in \mathbb{R}^{(c+1) \times \dots \times (c+1) \times c}$ and $\bar{v} = [v; 1]$ represents the homogeneous coordinate of $v$. The number of parameters required by the full tensor aggregation function are $O(c(c+1)^L)$.

\subparagraph{Hosvd-LSTM} \cite{Castellana2020ECAI} relies on the High Order Singular Value Decomposition (HOSVD) \cite{Kolda2009} of the tensor defined in the Full-LSTM. The aggregation function is:
\begin{equation}
    f(h_{v1},\dots,h_{vL}) = q\left(g^{\Tensor{G}}\left(u_1(h_{v1}), \dots, u_L(h_{vL}\right)\right),
    \label{eq:hosvdLSTM}
\end{equation}
where $u_1, \dots, u_L: \mathbb{R}^c \rightarrow \mathbb{R}^r$ and $q: \mathbb{R}^r \rightarrow \mathbb{R}^c$ are affine maps, while $g^{\Tensor{G}}: \mathbb{R}^r \times \dots \times \mathbb{R}^r \rightarrow \mathbb{R}^r$ is a multi-affine map defined through the tensor \Tensor{G}. The elements $\{u_1, \dots, u_L, q\}$ and  $g^{\Tensor{G}}$ are the equivalent, respectively, of the \emph{mode matrices} and the \emph{core tensor} of the decomposition. $r$ is the decomposition rank and we assume it is equal for each mode. The number of parameters required by the HOSVD aggregation function are $O(Lcr + r(r+1)^L)$. 

\subparagraph{Canonical-LSTM} exploits the canonical decomposition \cite{Kolda2009} of the tensor defined in the Full-LSTM. The aggregation function is:
\begin{equation}
    f(h_{v1},\dots,h_{vL}) = q\left(u_1(h_{v1})\odot \dots \odot _L(h_{vL})\right),
    \label{eq:canonicalLSTM}
\end{equation}
where  $u_1, \dots, u_L: \mathbb{R}^c \rightarrow \mathbb{R}^r$ and $q: \mathbb{R}^r \rightarrow \mathbb{R}^c$ are affine maps. The elements $\{u_1, \dots, u_L, q\}$ are the equivalent of the \emph{factor matrices} of the decomposition. The number of parameters required by the canonical aggregation function are $O(Lcr)$, where $r$ is the decomposition rank.

\subparagraph{TT-LSTM} uses the Tensor Train (TT) decomposition \cite{Oseledets2011} of the tensor defined in the Full-LSTM. The aggregation function is defined as:
\begin{equation}
    f(h_{v1},\dots,h_{vL}) = q\left(u_L^{\Tensor{T}_L}(\dots (u_2^{\Tensor{T}_2}( u_1(h_{v1}), h_{v2}),\dots, h_{vL}) \right),
    \label{eq:ttLSTM}
\end{equation}
where $u_1: \mathbb{R}^c \rightarrow \mathbb{R}^r$ and $q: \mathbb{R}^r \rightarrow \mathbb{R}^c$ are affine maps, while $u_2^{\Tensor{T}_2}, \dots, u_L^{\Tensor{T}_L}: \mathbb{R}^c \times \mathbb{R}^r \rightarrow \mathbb{R}^r$ are multi-affine maps. The elements $\{u_1, u_2^{\Tensor{T}_2}, \dots, u_L^{\Tensor{T}_L}, q\}$ are the equivalent of the \emph{cores} of the decomposition. The number of parameters required by the tensor train aggregation function are $O(cr + Lcr^2)$, where $r$ is the decomposition rank that we assume equal for each mode.

\section{Experimental Analysis}\label{sec:exp_analysis}
 To assess the impact of the maximum output degree in tree-structured learning problems, we test the Tree-LSTM models introduced in Section \ref{sec:tree-LSTM} on two classification problems. The architecture consists of a Tree-LSTM model, with different tensor-based aggregators, encoding trees to a fixed size representation (i.e. the root hidden state) that is then fed to a classifier.  In the first task, the classifier is a simple linear layer; in the second one, is a two-layer neural network with $20$ hidden neuron for each layer. The output of the classifier is then fed to a softmax and used to compute loss function (i.e. the negative log-likelihood).

All values reported are averaged over three executions, to account for randomisation effects due to initialisation. All models are trained using AdaDelta algorithm \cite{zeiler2012adadelta} and therefore no learning rate is set; all weights are initialised using Kaimining normal function \cite{He2015}. The code can be found here \footnote{\url{https://github.com/danielecastellana22/tensor-tree-nn}}.

\subparagraph{Boolean Sentences.} \label{sec:bool}
The goal of this task is to predict the output of a sequence of operations on boolean values. Each sentence is represented with its syntax tree 
where $\{0,1\}$ are the truth values which appear only on leaves, and \emph{OR}, \emph{AND} , \emph{IMPLY} are the logical operators which appear only on internal nodes. The tree maximum outdegree $L$ is defined by the number of inputs for each operators: we build four different dataset setting $L=\{2,3,4,5\}$. The results when $L>2$ is obtained by folding the input list with the logical operator from left to right (i.e. the fold left reduction is applied). Each dataset contains $10$k trees: $7000$ in the training set, $1000$ in the validation set and $2000$ in the test set. The height of each tree is between $4$ and $8$.
\begin{figure}[t]
    \centering
    \begin{subfigure}{0.49\textwidth}
        \includegraphics[width=\textwidth]{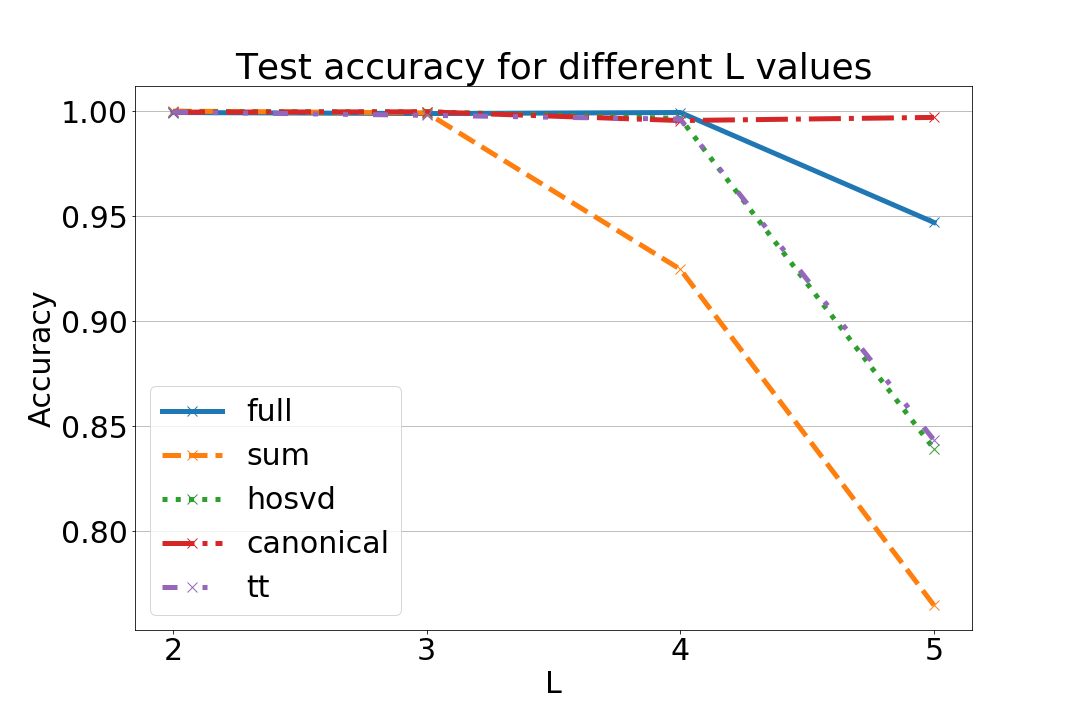}
        \caption{}
        \label{fig:bool_L}
    \end{subfigure}
    \begin{subfigure}{0.49\textwidth}
        \includegraphics[width=\textwidth]{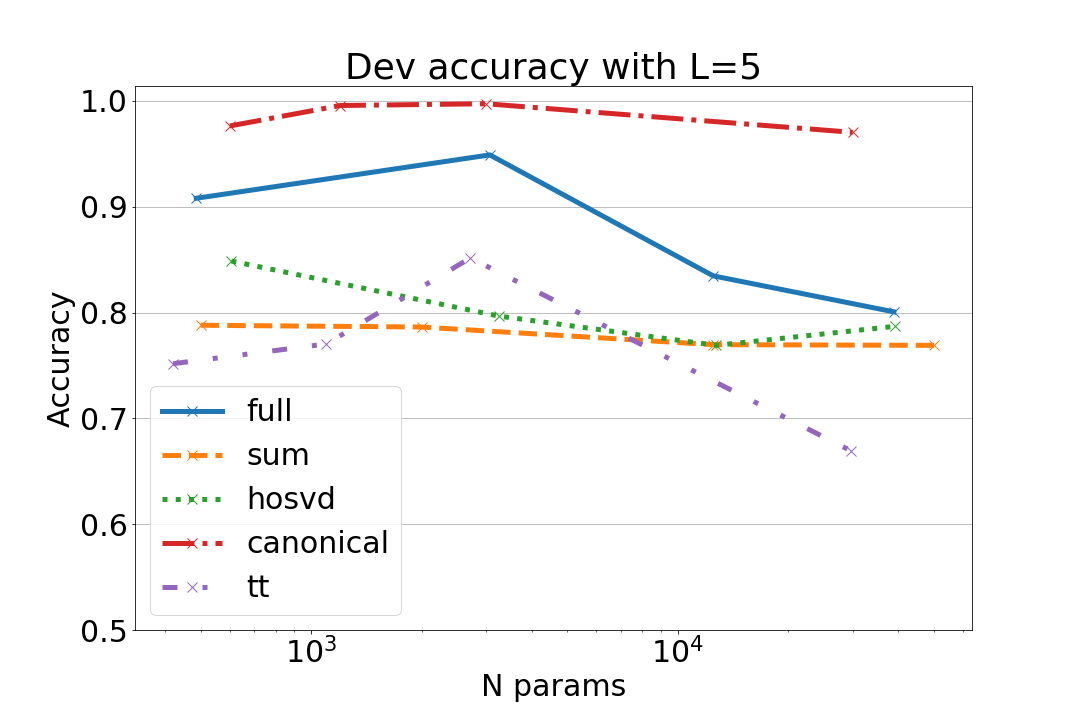}       
        \caption{}
        \label{fig:bool5_nparams}
    \end{subfigure}
    \caption{On the left, we report test results on different boolean dataset varying $L$. On the right, we report grid search results on boolean dataset with $L=5$.}
\end{figure}

In Fig. \ref{fig:bool_L}, we report the test results of each model on boolean datasets with different outdegree; for each dataset, the best models configuration is obtained through a grid search. When increasing the value of $L$, the task becomes more challenging due to the exponential growth of the possible interactions between the hidden child states. The results show that using a simple aggregation function such as the sum leads to a sub-optimal solution when $L>3$. The task with $L=5$ is challenging also for tensor models: however, the proposed Canonical-LSTM model can effectively solve the task.

In Figure \ref{fig:bool5_nparams}, we report the results of the grid search for each model on the most challenging task (i.e. $L=5$). It is interesting to highlight that the Canonical-LSTM outperforms all other models using less than $3$k parameters for each aggregation function.

\subparagraph{List Operations} \label{sec:ListOps}
The goal of this task is to predict the solution of a sequence of summary operations on lists of single-digit integers, written in prefix notation. An element in the dataset consists of a sequence of operations and its solution (which is also a single-digit integer). See \cite{Nangia2018} for more details on the dataset. Each sequence is represented by its syntax tree; the tree maximum outdegree is five (i.e. $L=5$) since all operations have at most five inputs. Also, the input digit $k$ is represented using a vector $x$ of size  $10$ which has the first $k+1$ entries equals to 1 and the other equals to 0 (e.g. if $k=2$, $x=[1,1,1,0,0,0,0,0,0,0]$). 

The dataset is already divided into training and test splits containing respectively 90\% and 10\% of the data \cite{Nangia2018}. We further sample 9\% of the training set to build a validation set. Hence, we obtain a training set which contains around 80k trees; validation and test set contains around 20k and 36k trees respectively.

\begin{figure}
    \centering
    \begin{subtable}{0.4\textwidth}
        \centering
        \scriptsize
        \begin{tabular}{c|c}
                Model & Test Accuracy\\
                \hline
                \hline
                Full & $82.02$ $(0.65)$ \\a
                Sum & $83.01$ $(1.06)$ \\
                Hosvd & $94.26$ $(0.48)$ \\
                Canonical & $\mathbf{95.82}$ $(1.11)$ \\
                TT & $95.47$ $(0.74)$ \\
        \end{tabular}
        \caption{}
    \end{subtable}
    \begin{subfigure}{0.59\textwidth}
        \centering
        \includegraphics[width=\textwidth]{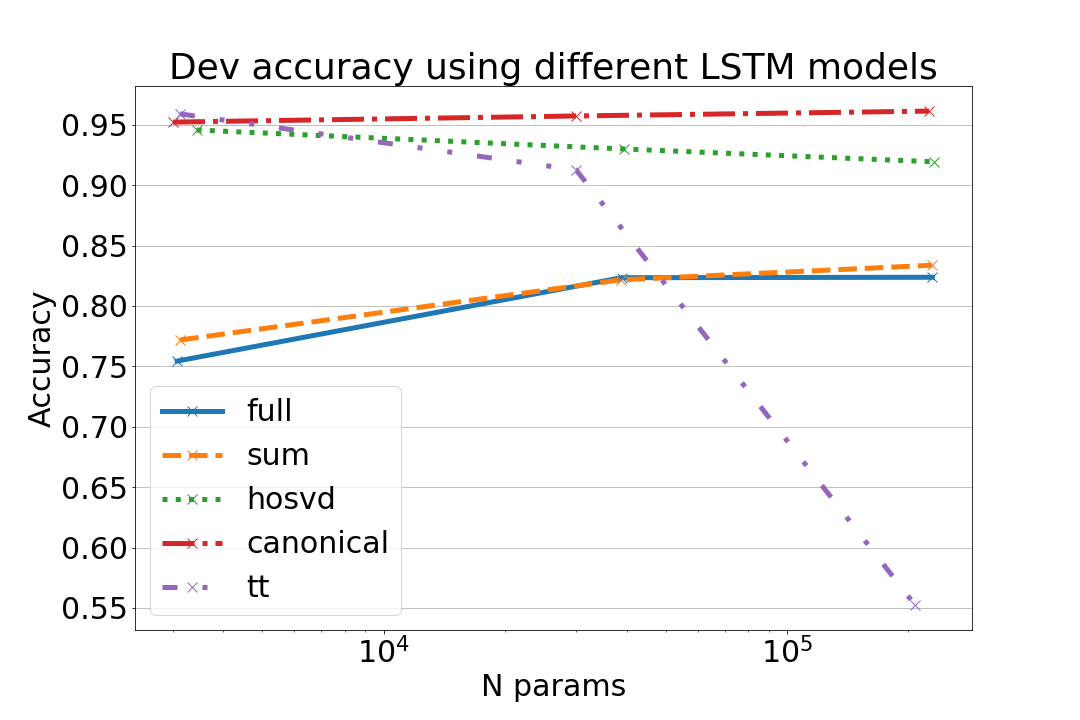}
        \caption{}
        \label{fig:listOps_nparams}
    \end{subfigure}
    \caption{We report test and grid search results on List-Ops dataset on the left and on the right respectively.}
\end{figure}
Figure \ref{fig:listOps_nparams} reports the results obtained in grid search by the models. For each configuration, we report the validation accuracy against the number of parameters required by each aggregation function. Again, Sum-LSTM is not able to reach satisfactory performance due to its simple aggregation function. Also  Full-LSTM obtains poor results on the task, most likely due to the small hidden state size (forced by otherwise intractable parameterisation). The Canonical-LSTM and Hosvd-LSTM outperform other models, reaching higher accuracy even with small parametrisation (less than $3$k parameters). The TT-LSTM performance degrades heavily when increasing the number of parameters: a further investigation is required to analyse this behaviour which is manifest also in the boolean dataset.

\section{Conclusion}\label{sec:conclusion}
In this paper, we have introduced new Tree-LSTM based models which leverage tensor decomposition to aggregate nodes context. The new aggregation functions may be seen as a trade-off between too simple function (such as the sum) and too complex maps (such as the full-tensor application). The advantages introduced by tensor decompositions are clearer when the tree maximum output degree $L$ increases.  When $L$ is higher, sum-based functions are not able to satisfactorily capture interaction among child information; on the contrary, full-tensor based functions cannot be even represented due to the curse of dimensionality. Our results encourage us to further study tensor decompositions in structured-data context to (i) understand the different decomposition biases, and (ii) to apply these powerful models on more complex task including analysing natural and artificial language data.


\begin{footnotesize}


\bibliographystyle{unsrt}
\bibliography{unpublished}

\end{footnotesize}


\end{document}